\documentclass[10pt,twocolumn,letterpaper]{article}

\usepackage{wacv}
\usepackage{times}
\usepackage{epsfig}
\usepackage{graphicx}
\usepackage{amsmath}
\usepackage{amssymb}
\usepackage{multirow}
\usepackage{url}

\def\wacvPaperID{47} 

\wacvfinalcopy 

\begin{document}

\title{CPM R-CNN: Calibrating Point-guided Misalignment in Object Detection}

\author{Bin Zhu, Qing Song, Lu Yang, Zhihui Wang, Chun Liu, Mengjie Hu\\
Pattern Recognition and Intelligent Vision Lab, Beijing University of Posts and Telecommunications\\
{\tt\small \{zhu\_bin, priv\}@bupt.edu.cn}
}

\maketitle
\thispagestyle{empty}

\begin{abstract}
   In object detection, offset-guided and point-guided regression dominate anchor-based and anchor-free method separately. Recently, point-guided approach is introduced to anchor-based method. However, we observe points predicted by this way are misaligned with matched region of proposals and score of localization, causing a notable gap in performance. In this paper, we propose CPM R-CNN which contains three efficient modules to optimize anchor-based point-guided method. According to sufficient evaluations on the COCO dataset, CPM R-CNN is demonstrated efficient to improve the localization accuracy by calibrating mentioned misalignment. Compared with Faster R-CNN and Grid R-CNN based on ResNet-101 with FPN, our approach can substantially improve detection mAP by 3.3\% and 1.5\% respectively without whistles and bells. Moreover, our best model achieves improvement by a large margin to 49.9\% on COCO test-dev. Code is available at \url{https://github.com/zhubinQAQ/CPM-R-CNN}.
\end{abstract}

\section{Introduction}

Object detection is one of the most fundamental research topics in computer vision, and many high-performing object detectors based on deep convolutional neural networks (CNN) have been proposed in recent years. These detectors are designed to solve object classification and localization problems, which can be generally divided into anchor-free methods and anchor-based methods. The anchor-free detectors\cite{Law2018CornerNet,tian2019fcos,Duan2019CenterNet} eliminate hyper parameters related to anchors, and the anchor-based detectors continuously optimize the offset based on the candidate boxes, which are the predefined anchors in one-stage methods \cite{Lin2017Focal,Liu2016SSD,Redmon2017YOLO9000,Zhang2018Single} or the candidate proposals in two-stage methods \cite{Ren2015Faster,He2017Mask,Cai2018Cascade}.

\begin{figure}
\begin{center}
\includegraphics[height=6cm]{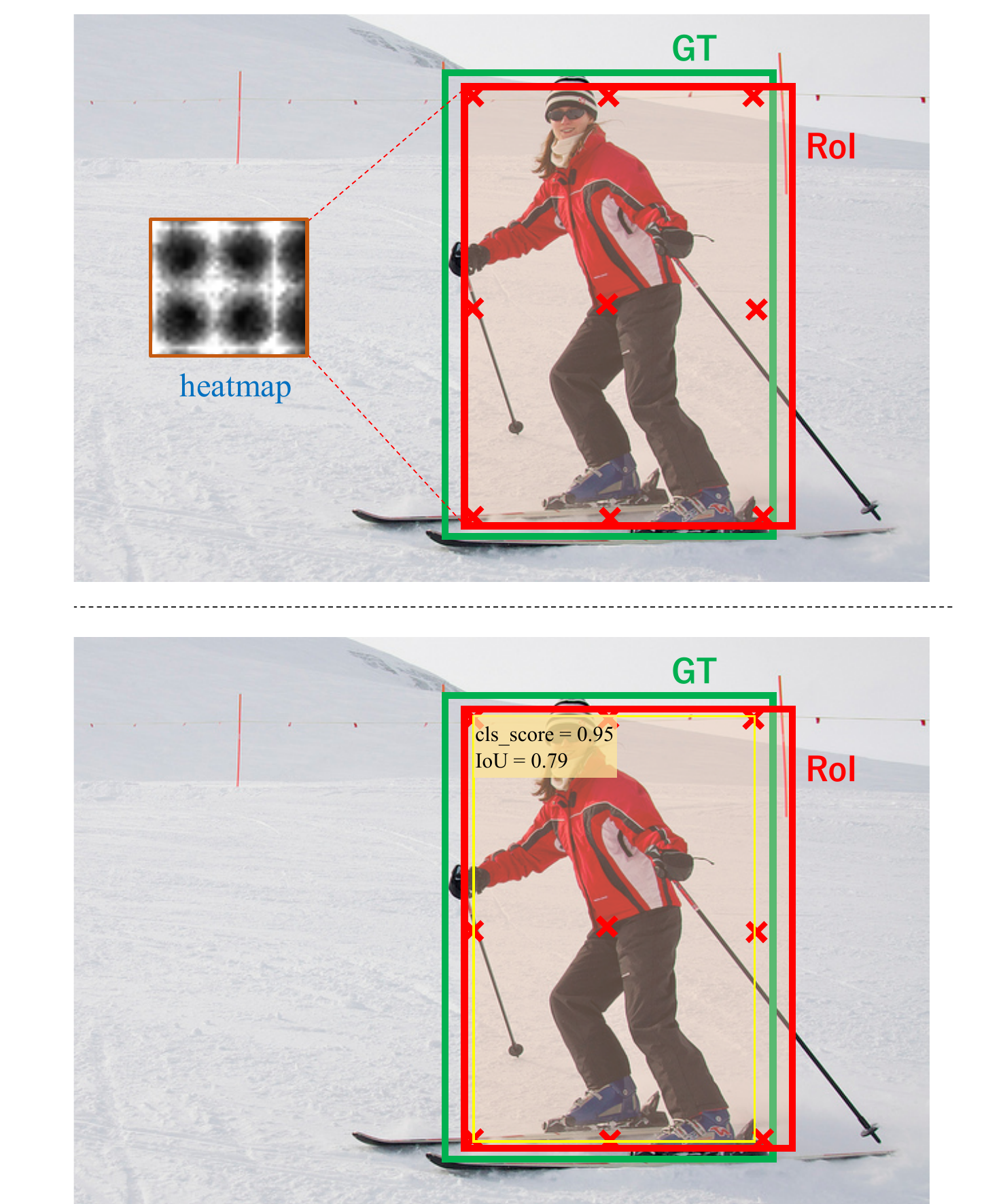}
\end{center}
   \caption{Illustration of misalignment in anchor-based point-guided method. Top: Heatmap prediction is limited by the distribution of RoI. Down: The quality of final predicted box (yellow) is not aligned with classification confidence of RoI (red) during inference.}
\label{fig:introduction}
\end{figure}

Mainstream two-stage anchor-based detection pipelines contain a large number of boxes (anchors, region of proposals, predicted bounding boxes). The whole localization regression process can be summarized as step-by-step spatial coordinate correction by offset predictions. Unlike these methods, Grid R-CNN~\cite{lu2019grid} provides a new idea to optimize the entire box stream. In the second stage, It converts original offset regression to the unique point-guided regression for localization, and it achieves similar performance with other two-stage method in this way. However, we observe that this point-guided method exists two common misalignment problems as shown in Fig.~\ref{fig:introduction}. Firstly, to predict points, major networks will make heatmaps by taking use of corresponding features of CNN. But for anchor-based architecture, it causes the misalignment between predicted points and matched RoIs. The reason lies in the way of generating features, based on location of proposals, it pays more attention inside the RoI. Thus it is inappropriate to regress the heatmap connected with each boundary point. Moreover, owing to the artificial regression starting status, anchor-based method limit the distribution of box stream, leading the distribution of proposals incomplete. The another misalignment problem is related to the score of localization. Actually, bounding box composed with predicted points is still shared with the classification confidence, but it is unwise, and the reasons can be ascribed to two aspects: (1) classification task is not sensitive to spatial informations, thus classification confidence is not aware of the actual quality of boxes. (2) there exists an inevitable hysteresis between the classification and localization regression, which means the obtained classification confidence is aligned with region of proposals rather the final boxes. If these boxes are not properly scored during inference, it might be wrongly regarded as false positive or false negative, resulting in a decrease of the detector performance.

In this work, we propose CPM R-CNN, a new framework for object detection to calibrate these misalignment in point-guided method. To overcome the misalignment between predicted points and matched RoIs, we design a cascade mapping module (CMM), applying extended region mapping~\cite{lu2019grid} in each stage. Guided by staged changing expansion ratio among the box stream, boxes can obtain more complete information out of the border. On the other side, in each cascade stage, variable proposals improve the diversity of matching with the target object, which can be regarded as enriching the distribution of box stream. Furthermore, for score of localization, we propose a fused scoring network to monitor the quality of the bounding box more accurately. It contains two modules, one is designed for monitoring the box spatial quality by learning the IoU score, called IoU Scoring Module (ISM), and the other is designed for obtaining a more accurate classification score, called Resampling Scoring Module (RSM). The final score of bounding box is formulated with outputs of these modules, and they are incorporated to eliminate misalignment between localization score and quality.

In extensive experiments on MS COCO datasets~\cite{Lin2014Microsoft}, our framework has significant performance over other point-guided methods. For example, compared with Grid R-CNN~\cite{lu2019grid} with a backbone of ResNeXt-101~\cite{Xie2017Aggregated} with FPN~\cite{Lin2017Feature}, our best model improve overall mAP by 49.9\% on COCO test-dev. In addition, it's worth noting that our detector has a consistent improvement on the large object by applying proposed CMM, achieving a 1.4\% AP$_{l}$ gain based on ResNet-50~\cite{He2016Deep} with FPN. Besides, CPM R-CNN is effective on higher IoU thresholds, for example, we get 13.4\% and 3.5\% AP improvement over Faster R-CNN and Grid R-CNN respectively at 0.9 IoU threshold. 

Our main contributions are as follows:
\begin{itemize}
\item  We point out the reason for bottleneck of detection accuracy in point-guided network lies in two kinds of misalignment problems. And we present CPM R-CNN to address these problems. 

\item In order to obtain more complete distribution of boxes, we propose a new approach called cascade mapping, which is proved to be effective according to experimental results.
    
\item We design a simple and effective fused scoring structure to correct the inappropriate and misaligned score of localization, and this wise scoring strategy bring great improvement over original scoring way in our ablation research.

\end{itemize}

\section{Related work}

Our new approach is based on two-stage object detection method, which falls into anchor-based object detection architectures. In addition, there are some researches about the misalignment of localization score in recent years, and we briefly review some related works as follows.

\subsection{Anchor-based Object Detection Architectures}

General state-of-the-art anchor-based object detectors can be divided into one-stage and two-stage detectors. Due to R-CNN architecture \cite{girshick2014rich}, two-stage detectors develop quickly in these years. Thereafter, Fast R-CNN~\cite{girshick2015fast} and SPP-Net~\cite{he2015spatial} extracted feature by adopting region-wise strategy to reduce redundant computing burden, sharing feature computations. Then in Faster R-CNN~\cite{Ren2015Faster}, the propose of region proposal network (RPN) achieved acceleration and end-to-end training. After the appearance of YOLOv2~\cite{Redmon2017YOLO9000} and SSD~\cite{Liu2016SSD}, the one-stage detectors show their great advantage on the computational efficiency. YOLOv2 designed an efficient backbone network to enable real-time detection. SSD used  multiple feature maps of multi-scale layers, enabling anchors assigned to objects at various scales. Although, there is still large room to improve for anchor-based methods. 

\begin{figure*}
	\centering
	\includegraphics[height=5.0cm]{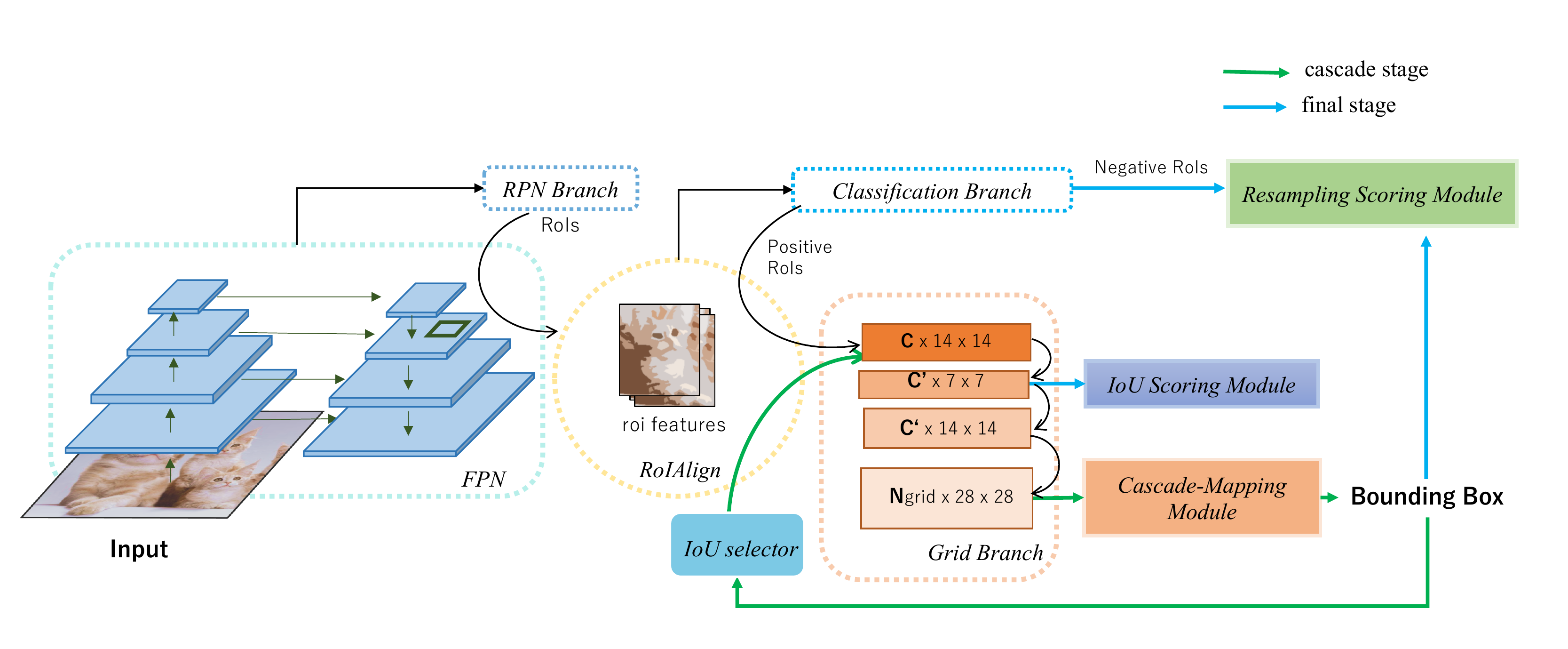}
	\caption{The illustration of CPM R-CNN. Given an input image, we employ FPN as the backbone architecture to generate features for RPN and RoIAlign. The RoI features from RoIAlign are fed into classification branch to classify RoIs. The positive RoI features are selected to regress the bounding box in grid branch, and boxes of higher IoU quality are chosen for the next stage. During the cascade forwarding process, we propose Cascade-Mapping Module (CMM), Resampling Scoring Module (RSM) and IoU Scoring Module (ISM) three modules to boost network performance. The green arrows represent the whole cascade operations and the blue arrows represent operations are only applied in the final stage. Details of CMM, RSM and ISM are shown in Fig.~\ref{fig:cmm} and Fig.~\ref{fig:score}.}
	\label{fig:pipeline}
\end{figure*}

Recently, to boost performance, new ideas and algorithms are proposed in plenty of works.~\cite{yang2018attention} designed a more powerful backbone. R-FCN~\cite{Dai2016R} modified original fully connected network in Faster R-CNN to achieve further computational reduction. RetinaNet~\cite{Lin2017Focal} noticed class imbalance in object detection, and solving the problem by using focal loss, bridging the performance gap between one-stage and two-stage detectors.~\cite{He2017Mask,yang2019parsing} enriched supervising clue by adding new task branch. RefineDet~\cite{Zhang2018Single} put two-step bounding regression into one-stage framework, generating more accurate refined anchors to improve detector performance. Cascade R-CNN~\cite{Cai2018Cascade} reformed the traditional cascade connection by taking IoU thresholds as limitation when proposals forwarding among all stages. Hence,  the output of each stage in Cascade R-CNN got optimized corresponding with their different distribution. Based on two-stage structure, Grid R-CNN~\cite{lu2019grid} introduced a grid guided localization mechanism to design the second stage. It focuses on  interrelationship of internal points. Specifically, by fusing spatial information of each grid, it achieved very similar performance with regression based method. 

Different from Grid R-CNN, CPM R-CNN focuses on stage-by-stage distribution optimization, thus our localization branch is designed more simplified and efficient and optimized by introducing a new cascade architecture. Moreover, based on unique coarse-to-fine mapping strategy, our approach is more appropriate for heatmap prediction.

\subsection{Localization Quality Correlation}
For object detection, it has been proved that the correlation between classification confidence and localization quality is not strong in many previous works. SoftNMS~\cite{Bodla2017Soft} used IoU to sieving low-quality boxes. In~\cite{Tychsen2018Improving}, DeNet~\cite{Tychsen2017DeNet} obtained a significantly improvement by applying Fitness NMS. Unlike SoftNMS, the localization quality in Fitness NMS was classified into multiple levels, and IoU prediction more like a classification task. IoU-Net~\cite{Jiang2018Acquisition} added a new IoU prediction branch to Faster R-CNN~\cite{Ren2015Faster}, regressing IoU score to get accurate rank in NMS. Thereafter, MS R-CNN~\cite{Huang2019Mask} transferred the misalignment problem to the mask task. Based on Mask R-CNN~\cite{He2017Mask}, it abandoned classification confidence for directly measuring mask quality, and proposed a MaskIoU head to regress the mask IoU during inference. 

Compared with IoU-Net, CPM R-CNN is a grid-guided method, therefore our IoU scoring module shares same heatmap input with grid branch. Meanwhile, we point out the weakness of IoU-based scoring method and propose resampling scoring module to optimize it's performance.

\section{Method}
In this part, we will introduce the architecture of CPM R-CNN. As is shown in Fig.~\ref{fig:pipeline}, our algorithm contains three major components: (1) Cascade grid guided localization branch predicts bounding boxes for the whole input. (2) CMM helps the regression loss function to converge stage by stage. (3) RSM and ISM work together to generate score for each box.

\subsection{Network Architecture}
CPM R-CNN adopts FPN~\cite{Lin2017Feature} as the backbone architecture and adds a  standard classification branch of Faster R-CNN~\cite{Ren2015Faster}. For localization, guided by~\cite{lu2019gridPlus}, we design our grid branch by grid guided regression instead of offset regression. It consists of downsampling and upsampling layers. And then the generated 28$\times$28 heatmap is delivered to CMM for regressing the bounding box. After that, IoU selector collects boxes with higher IoU to the next cascade stage. In the final stage, for corresponding score prediction, ISM gets 7$\times$7 feature maps from grid branch, and RSM also receives negative and positive samples from classification branch and CMM separately. The detail architecture of CMM, ISM and RSM are illustrated in the following sections.

\begin{figure}
	\centering
	\includegraphics[height=5.2cm]{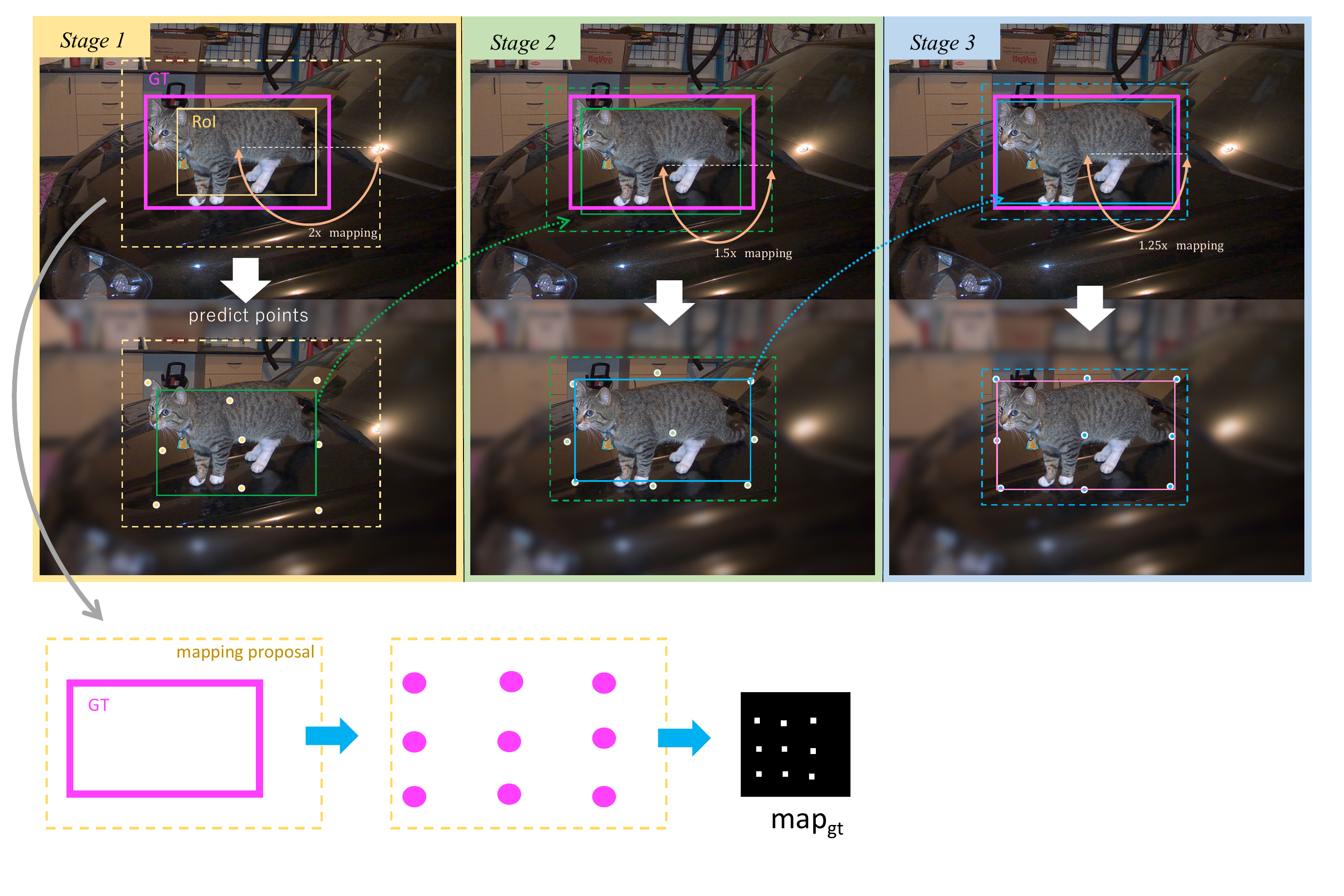}
	\caption{The pipeline of CMM. Different mapping ratios are applied in the mapping operation of each stage. Boxes will be adjusted more and more accurately by our coarse-to-fine strategy. }
	\label{fig:cmm}
\end{figure}

\subsection{Cascade Mapping Strategy}
As a two-stage detector, the RoI feature is the basis for following feature extraction in second-stage subbranch. Especially in most point-guided methods, final points directly rely on heatmap prediction which can be nearly regarded as pixel-level classification task, thus it requires RoI feature to consist more complete and accurate information than traditional offset-guided method. Therefore, to address this, we expect that the localization of proposals is adequately close to corresponding ground truth. But RPN is limited by artificial superparameters of anchors. It leads the misalignment between matched proposals and second-stage regression. 

Since network lacks aligned proposals, we introduce cascade mapping module (CMM) to improve this circumstances. As shown in Fig.~\ref{fig:cmm}, we enrich the proposal by two steps. Firstly, we apply mapping operation $M_{j}$ on grid branch $g_{j}$ in each stage, which means the proposals are expanded with a specific mapping ratio after RoIAlign operation to match ground truth. Then, matched boxes are converted to heatmap $map_{gt}$ for loss regression, and the coordinates of predicted points will also be mapped back to the original image to formulate next proposal $B'$ in return. To get proposals with high quality, then we assign coarse-to-fine mapping ratios and IoU thresholds to cascade stages. Given a feature $P_{i}$ generated from $i$-th scale FPN branch and the proposals $B_{j}$ from the $j$-th cascade stage, the whole process can be formulated as follows. 

\begin{equation}
B' = S_{j}(M_{j}(g_{j}(RA(P_{i}, B_{j})))
\label{eq:cm1} 
\end{equation}
where $B_j\in \mathbb{R}_j$ and $\mathbb{R}_j$ indicates all reserved box selected by $S_{j}$ with corresponding IoU, $RA$ means RoIAlign operation, $g_{j}$ represents grid branch shown in  Fig.~\ref{fig:pipeline}. Thereafter, we define loss function of CMM as:
\begin{equation}
\mathcal{L}_{CMM} = \sum_{j\in N}\beta_{j}\omega\mathcal{L}_{BCE}(g_{j}, map_{gt})
\label{eq:cm3} 
\end{equation}
where $N$ equals total number of stages, $\beta_{j}$ indicates changing loss weights in each stage and $\omega$ represents fixed loss weight of grid branch. 

The cascade structure with especial cascade mapping strategy of CPM is to assist grids get proper feature for effective grid-guided regression, so as to precisely predict the final heatmap. As for cascade offset-guided regression like Cascade R-CNN, it improves the IoU distribution of samples to predict the offset of boxes more accurately. 

\subsection{Fused Scoring Network}
Considering the misalignment between classification score and practical location quality, we introduce more proper supervision in cascade grid branch for getting more complete spatial information. From another perspective, the refined detections are also used to optimize classification by updating sampling space. To this end, two scoring modules are proposed, namely IoU scoring module (ISM) and resampling scoring module (RSM).

\begin{figure*}
	\centering
	\includegraphics[height=3.9cm]{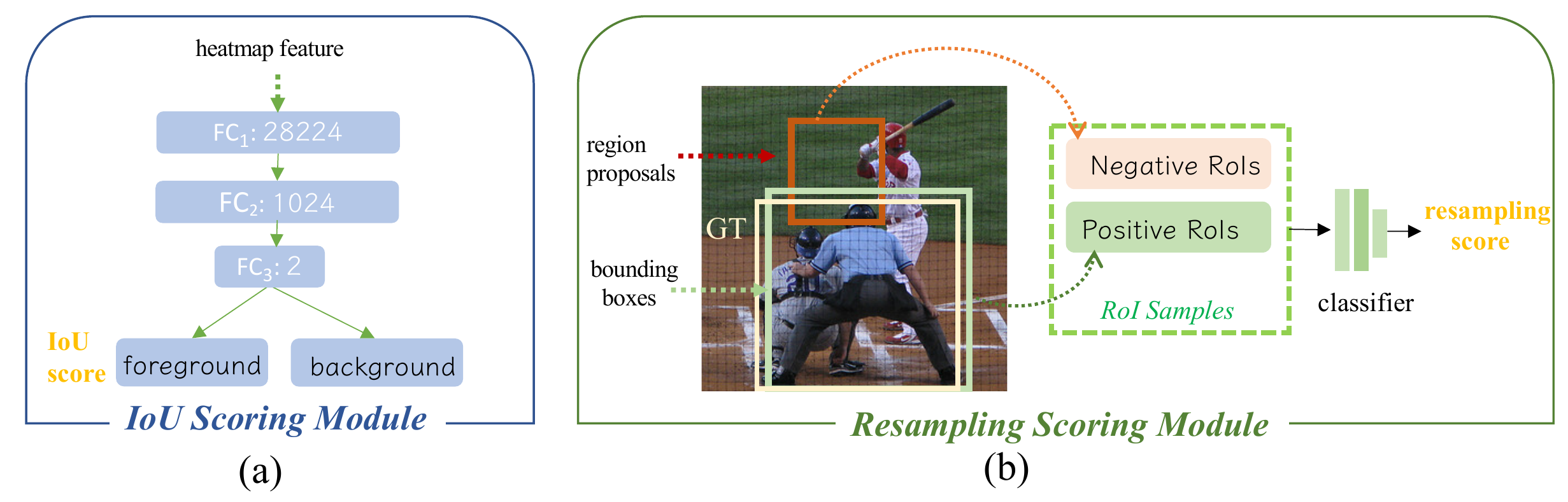}
	\caption{(a) : The architecture of IoU Scoring Module. (b) : The architecture of Resampling Scoring Module.}
	\label{fig:score}
\end{figure*}

\subsubsection{IoU Scoring Module}
For a classic two-stage network architecture, the localization branch focuses on coordinate regression. This means the spatial localization quality of each predicted box is optimized and measured by calculating offset during training stage. Similarly, we believe that predicted bounding boxes also require an index to measure their spatial quality in inference stage. Therefore, we propose IoU Scoring Module (ISM). As shown in Fig.~\ref{fig:score} (a), it only consists of three simple fully connected layers but proved efficient. To gain computational reduction by decreasing input channels, ISM is especially applied in the final cascade stage like Fig.~\ref{fig:pipeline} shows. With the supervision of proposals actual IoU, ISM will generate scores of two classes (foreground and background) for each proposal by utilizing feature with minimum scale in grid branch. In detail, this module can be formulated as:
\begin{equation}
score_{ISM} = f(g'(RA(P_{i}, B_{f})))
\end{equation}
where $f$ represents a simple classification function, $g'$ denotes $g$ in Equ.~\ref{eq:cm1} without final 4$\times$ upsampling, and $B_{f}$ means boxes generated from final stage.

\subsubsection{Resampling Scoring Module}\label{section:rsm}
With introducing predicted IoU to formulate localization score in ISM, detector performance is almost aligned with actual localization quality. However, this strategy is defective. Specifically, the objects in our training dataset can be divided into two categories: (1) complete object, which means one image contains all parts of the target object. (2) incomplete object, which means only a part of the object is in the image. Obviously, In consideration of characteristics of different scales, small-scale object is easily classified as first type, and large-scale object is another one. When ISM uses calculated IoU for training, it obtains ability of perceiving the integrity of objects. This benefits  getting localization quality of complete objects. Meanwhile, it is not friendly to incomplete objects, and this shortage can be found in Fig.~\ref{fig:rsm}. It prefers to predict IoU  between bounding box with overall appearance in actual world rather with ground truth in an image.

To remedy this, we propose a resampling scoring module (RSM) as shown in Fig.~\ref{fig:score} (b).  And it's worth noting that this is core difference among other IoU strategy like IoU Net. As for input samples,  the negative RoIs are generated from original RPN, and the positive RoIs are entire bounding boxes produced from cascade grid branch in the final stage. Then we pass them to the classifier. The purpose of our design is to align the classification quality with the final bounding boxes predicted by the detector during inference stage. RSM provides a new classification confidence to complement ISM. $\gamma$ is a balanced factor for the final score prediction. Thus the final fused score of localization is defined as:
\begin{equation}
score_{fused} = (score_{cls} \times score_{ISM}) ^ {\gamma} \times score_{RSM} ^ {(1-\gamma)}\label{eq:fusedscore} 
\end{equation}

 Moreover, different from normal IoU prediction, our scoring network is designed to concern how to strengthen the connection between localization quality and classification confidence rather than only introducing a new index to measure localization. By introducing a simple design of RSM, we also fix its weakness showed in Fig.~\ref{fig:rsm}. Thus in our proposed fused scoring network, both localization quality and classification quality all properly participate in final score computation.
 
\subsection{Implementation Details}\label{section:set}
\subsubsection{Network Setting}
We use ResNet-50/101~\cite{He2016Deep} based FPN~\cite{Lin2017Feature} as the backbone network. RPN~\cite{Ren2015Faster} is adopted to generate region of proposals by setting 256 anchors with same negative and positive sample ratio, and the IoU threshold is 0.3 and 0.7 respectively. For COCO dataset~\cite{Lin2014Microsoft}, all the input images are resized to have 800 pixels along the short axis and a maximum of 1333 pixels along the long axis. We apply RoIAlign~\cite{He2017Mask} to generate RoI features and set the output resolution 7 in classification branch and 14 in cascade grid branch. For classification branch, we samples 512 RoIs composed with 3/4 negative samples and 1/4 positive samples, and set threshold as 0.5 to separate them. In cascade grid branch, positive samples are selected with IoU thresholds\{0.5, 0.6, 0.7\} separately for training.  

\subsubsection{Optimization}
To balance training processes, our network is optimized via a joint loss function $\mathcal{L}_{all}$. $\mathcal{L}_{RSM}$ has same loss function with $\mathcal{L}_{cls}$, and $\mathcal{L}_{CMM}$, $\mathcal{L}_{ISM}$ are defined in Equ.~\ref{eq:cm3} and Equ.~\ref{eq:ism} respectively.
\begin{equation}
\mathcal{L}_{all} = \lambda_{1}\mathcal{L}_{rpn} + \lambda_{2}\mathcal{L}_{cls} + \lambda_{3}\mathcal{L}_{scoring} + \lambda_{4}\mathcal{L}_{CMM}
\end{equation}
\begin{equation}
\mathcal{L}_{scoring} = \alpha_{1}\mathcal{L}_{RSM} + \alpha_{2}\mathcal{L}_{ISM}
\end{equation}

In the experiments, $\lambda_{1}$ to $\lambda_{4}$ and $\alpha_{1}$, $\alpha_{2}$ are all set to 1,  and we set $\beta_{1}$ to 1 in Equ.~\ref{eq:cm3}, then decrease it a half in the next stage. We adopt stochastic gradient descent (SGD) with 4e-5 weight decay and 0.9 momentum for optimizing the  $\mathcal{L}_{all}$. The backbone is pretrained on ImageNet dataset~\cite{russakovsky2015imagenet}, and the rest of parameters are initialized by~\cite{He2015Delving}. We use 8 Nvidia TITAN Xp GPUs to train models for 180K (2$\times$) iterations, and each mini-batch has 2 images per GPU. The learning rate is warmed up~\cite{Goyal2017Accurate} and initialized with 0.02, then divided by 10 in the 120K and 160K iterations respectively. Input images are processed with horizontally flipping without other data augmentations during training.

\subsubsection{Inference}
RPN produces 1000 RoIs for RoIAlign and classification when FPN is adopted in the inference stage. Thereafter, we apply NMS with 0.3 IOU threshold to select satisfied RoIs which is 96 at most per image. Then the location of these RoIs are corrected precisely stage by stage in cascade grid branch. Among this process, classification confidence, IoU score and resampling score are produced from classification branch, ISM and RSM respectively, and they jointly construct Equ.~\ref{eq:fusedscore} to calculate the $score_{fused}$. $\gamma$ is set to 0.8 in this equation. After that the fused score is applied to rank all detections for final AP computation.

\section{Experiments}
\subsection{Component-wise Analysis and Diagnosis}
In this section, we will gradually decompose our method for revealing the effect of each part, and all models in these experiments are trained on COCO dataset based ResNet-50-FPN backbone networks. In order to make a fair comparison, we adopt no trick or multi-scale data augmentations during the whole process when analyzing experiments. Compared with separate training method, our proposed CPM R-CNN achieve an absolute improvement of 1.4\% in AP as shown in Table \ref{table:ablation}. Compared with Faster R-CNN and Grid R-CNN based on ResNet-101 with FPN, our approach can substantially improve detection mAP by 3.3\% and 1.5\% respectively in Table~\ref{table:compare}.

\setlength{\tabcolsep}{3pt}
\begin{table}
\renewcommand\arraystretch{1.2}
\begin{center}
\begin{tabular}{ccc|c|ccccc}
\noalign{\bigskip}
\hline
CMM & ISM & RSM & AP & AP$_{0.5}$ & AP$_{0.75}$ & AP$_{S}$ & AP$_{M}$ & AP$_{L}$ \\
\hline
& &  & 39.9 & 58.8 & 42.8 & 22.3 & 43.1 & 53.3\\
\checkmark & &  &  40.7 & 58.8 & 43.3 & 22.5 & 44.1 & 54.5\\
& \checkmark &  & 40.5 & 58.1 & 43.5 & 22.4 & 43.9 & 54.2\\
& & \checkmark & 40.6 & 58.8 & 43.5 & 22.5 & 43.5 & 54.4\\
 &  \checkmark & \checkmark & 40.9 & 58.5 & 44.2 & 22.8 & 44.0 & 54.7\\
  \checkmark & & \checkmark & 40.9 & 58.8 & 43.6 & 22.4 & 44.1 & 54.9\\
 \checkmark & \checkmark & & 41.1 & 58.6 & 43.8 & 22.7 & 44.6 & 55.6\\
\checkmark & \checkmark & \checkmark & \bf41.3 & \bf59.0 & \bf44.1 & \bf23.1 & \bf44.8 & \bf55.7\\

\hline
\end{tabular}
\end{center}
\caption{
Comparison among different settings of AP quality on the COCO 2017 validation dataset. CMM has 2 cascade stages in this experiment. We consider the classification branch parallel with our basic grid branch as the baseline.
}
\label{table:ablation}
\end{table}

\setlength{\tabcolsep}{4pt}
\begin{table}
\renewcommand\arraystretch{1.2}
\begin{center}
\begin{tabular}{c|c|c|ccccc}
\hline
backbone & method & schedule & AP\\
\hline
\multirow{3}*{R-50 w FPN} & Faster R-CNN & 2$\times$ & 37.6\\
& Grid R-CNN & 2$\times$ & 40.3\\
& CPM R-CNN & 2$\times$ & \bf41.5\\
\hline
\multirow{3}*{R-101 w FPN} & Faster R-CNN & 2$\times$ & 39.9\\
& Grid R-CNN & 2$\times$ & 41.7\\
& CPM R-CNN & 2$\times$ & \bf43.2\\
\hline
\end{tabular}
\end{center}
\caption{
The detection performance of different methods on COCO $2017~val$. Grid R-CNN results come from mmdetection~\cite{mmdetection}. There are 3 stages in CPM R-CNN.}
\label{table:compare}
\end{table}

\setlength{\tabcolsep}{4pt}
\begin{table}
\renewcommand\arraystretch{1.2}
\begin{center}
\begin{tabular}{c|c|c|ccccc}
\hline
base method & stage & impr & AP\\
\hline
\multirow{4}*{Cascade R-CNN} & 2 & none & 39.8\\
& 2 & +ISM & 39.9\\
& 2 & +RSM & 40.0\\
& 2 & +ISM\&RSM & \bf40.2\\
\hline
\end{tabular}
\end{center}
\caption{
The detection performance of applying different component of CPM on Cascade R-CNN. The backbone of Cascade R-CNN is ResNet-50.}
\label{table:cascade}
\end{table}

\subsubsection{Cascade Mapping Module}
Our proposed CMM assists the network in fine-tuning the proposals stage by stage. Thus the final predicted heatmap is gradually corrected to align with ground truth by this way. As presented in Table~\ref{table:ablation}, the network performs consistent 0.8\% gain in AP with calibrating this misalignment. In Table~\ref{table:stageperformance}, it shows stable improvement with increased stages, and the following stage in CMM achieves better performance at higher IoU thresholds. This can be resulted from continuous optimization of the IoU distribution among all stages. Meanwhile, it slightly reduces performance at lower IoU thresholds. Moreover, CMM with 3 stages achieves 0.4\%, 1.0\% and 1.4\% improvement over baseline on AP$_{S}$, AP$_{M}$ and AP$_{L}$, which also proves its effectiveness on the large objects. Moreover, in Table~\ref{table:mappingperformance}, it's worth noting that this improvement on large objects mostly comes from our reasonable design of mapping ratio rather than the IoU filtering mechanism of normal cascade structure. 

\setlength{\tabcolsep}{3pt}
\begin{table}
\small
\renewcommand\arraystretch{1.1}
\begin{center}
\begin{tabular}{c|c|c|ccc|ccccc}
\hline
stages & test stage & AP & AP$_{S}$ & AP$_{M}$ & AP$_{L}$ & AP$_{0.5}$ & AP$_{0.7}$ &  AP$_{0.9}$ \\
\hline
1 & baseline & 39.9 & 22.3 & 43.1 & 53.3 & 58.8 & \bf49.7 & 18.3\\
 \hline
\multirow{2}*2 & 1 & 39.9 & 22.3 & 43.3 & 53.2 & 58.8 & 49.7 & 18.5\\
& 2 & 40.7 & 22.5 & 44.1 & 54.5 & 58.8 & 47.6 & 21.1  \\
\hline
\multirow{3}*3 & 1 & 39.8 & 22.2 & 43.1 & 53.1 & \bf59.1 & 47.1 & 18.1\\
& 2 & 40.7 & 22.6 & 44.0 & 54.6 & 59.0 & 47.5 & 21.2\\
& 3  & \bf40.8 & \bf22.7 & \bf44.1 & \bf54.7 & 58.9 & 47.6 & \bf21.4\\
\hline
\end{tabular}
\end{center}
\caption{
The performance of CMM in each single stage. 
}
\label{table:stageperformance}
\end{table}

\setlength{\tabcolsep}{3pt}
\begin{table}
\small
\renewcommand\arraystretch{1.1}
\begin{center}
\begin{tabular}{c|c|c|cccccccc}
\hline
mapping ratio & test stage & AP & AP$_{S}$ & AP$_{M}$ & AP$_{L}$\\
\hline
\multirow{3}*{(2,2,2)} & 1 & 39.9 & 22.3 & 43.2 & 53.2\\
& 2 & 40.4 & 22.3 & 43.7 & 53.9\\
& 3  & 40.3 & 22.5 & 43.6 & 53.8\\
\hline
\multirow{3}*{(2,1.5,1.25)} & 1 & 39.8 & 22.2 & 43.1 & 53.1\\
& 2 & 40.7 & 22.6 & 44.0 & 54.6\\
& 3  & \bf40.8 & \bf22.7 & \bf44.1 & \bf54.7\\
\hline
\end{tabular}
\end{center}
\caption{
The effectiveness of mapping ratio in CMM. Model is effectively optimized by applying coarse-to-fine mapping ratio in each stage.
}
\label{table:mappingperformance}
\end{table}


\subsubsection{IoU Scoring Module} 
\begin{figure}
	\centering
	\includegraphics[height=4.5cm]{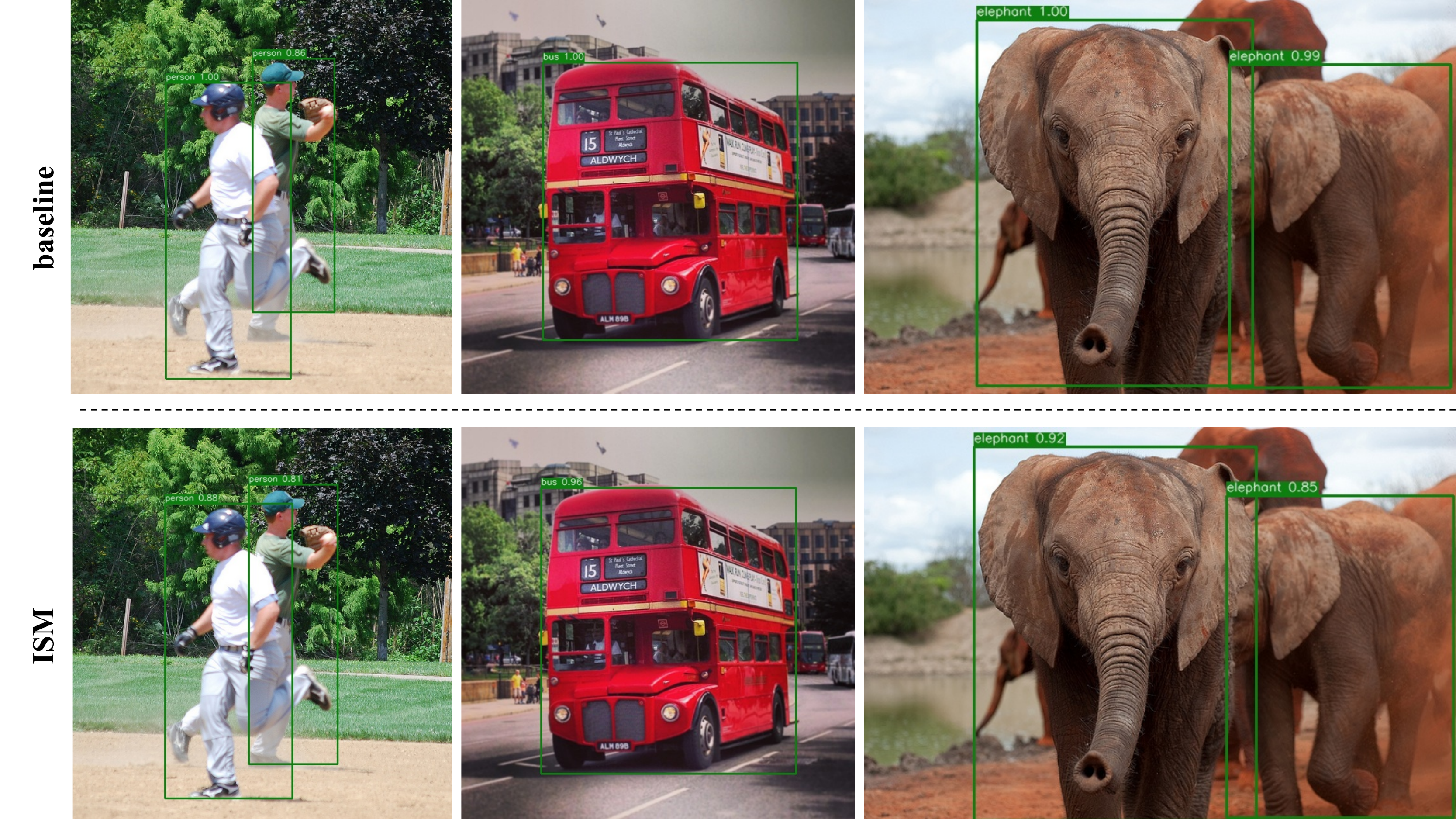}
	\caption{Qualitative comparison on COCO validation. Bounding box gets more proper score when ISM is applied. Models all come from Table~\ref{table:ablation}.}
	\label{fig:ism}
\end{figure}

To optimize our ISM, the loss function $\mathcal{L}_{ISM}$ is designed as:
\begin{equation}
\mathcal{L}_{ISM} = L(IoU_{predict}, IoU_{target})
\label{eq:ism}
\end{equation}
where $L \in (\ell_{1}, \ell_{2}, smooth~\ell_{1}~loss)$. Table~\ref{table:iouloss} shows that $\ell_{2}$ loss achieves the best performance, thus we adopt $\ell_{2}$ loss for all experiments. The results in Table \ref{table:ablation} also show that our proposed ISM independently brings 0.6\% absolute improvements in AP. We attribute this to the sensitiveness of IoU. It directly reflects the overlapping relationship with the ground truth. By applying accurate score to measure quality of spatial location, ISM eliminates misaligned confidence in classification as shown in Fig.~\ref{fig:ism}.

\setlength{\tabcolsep}{4pt}
\begin{table}
\renewcommand\arraystretch{1}
\begin{center}
\begin{tabular}{c|c|ccccc}
\noalign{\bigskip}
\hline
loss type & AP & AP$_{0.5}$ & AP$_{0.75}$ & AP$_{S}$ & AP$_{M}$ & AP$_{L}$\\
\hline
baseline & 39.9 & 58.8 & 42.8 & 22.3 & 43.1 & 53.3\\
$\ell_{1}$ & 40.0 & 57.8 & 43.0 & 22.2 & 43.4 & 53.8\\
$smooth~\ell_{1}$ & 40.3 & 58.0 & 43.3 & 22.4 & 43.6 & 54.3\\
$\bf\ell_{2}$ & \bf40.5 & \bf58.1 & \bf43.5 & \bf22.4 & \bf43.9 & \bf54.2\\
\hline
\end{tabular}
\end{center}
\caption{
The effectiveness of different loss type on COCO $2017~val$.
}
\label{table:iouloss}
\end{table}

\subsubsection{Resampling Scoring Module}
\begin{figure}
	\centering
	\includegraphics[height=4cm]{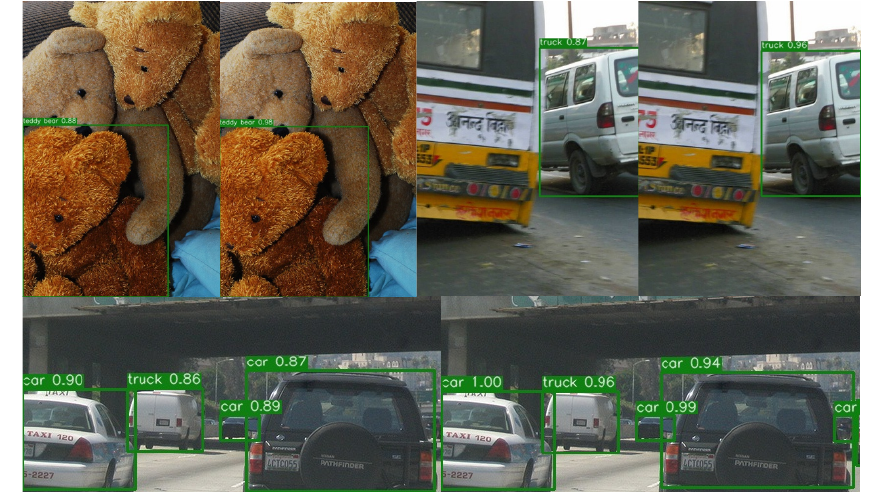}
	\caption{Examples of applying ISM (left) and RSM (right). }
	\label{fig:rsm}
\end{figure}

\begin{figure}
	\centering
	\includegraphics[height=4cm]{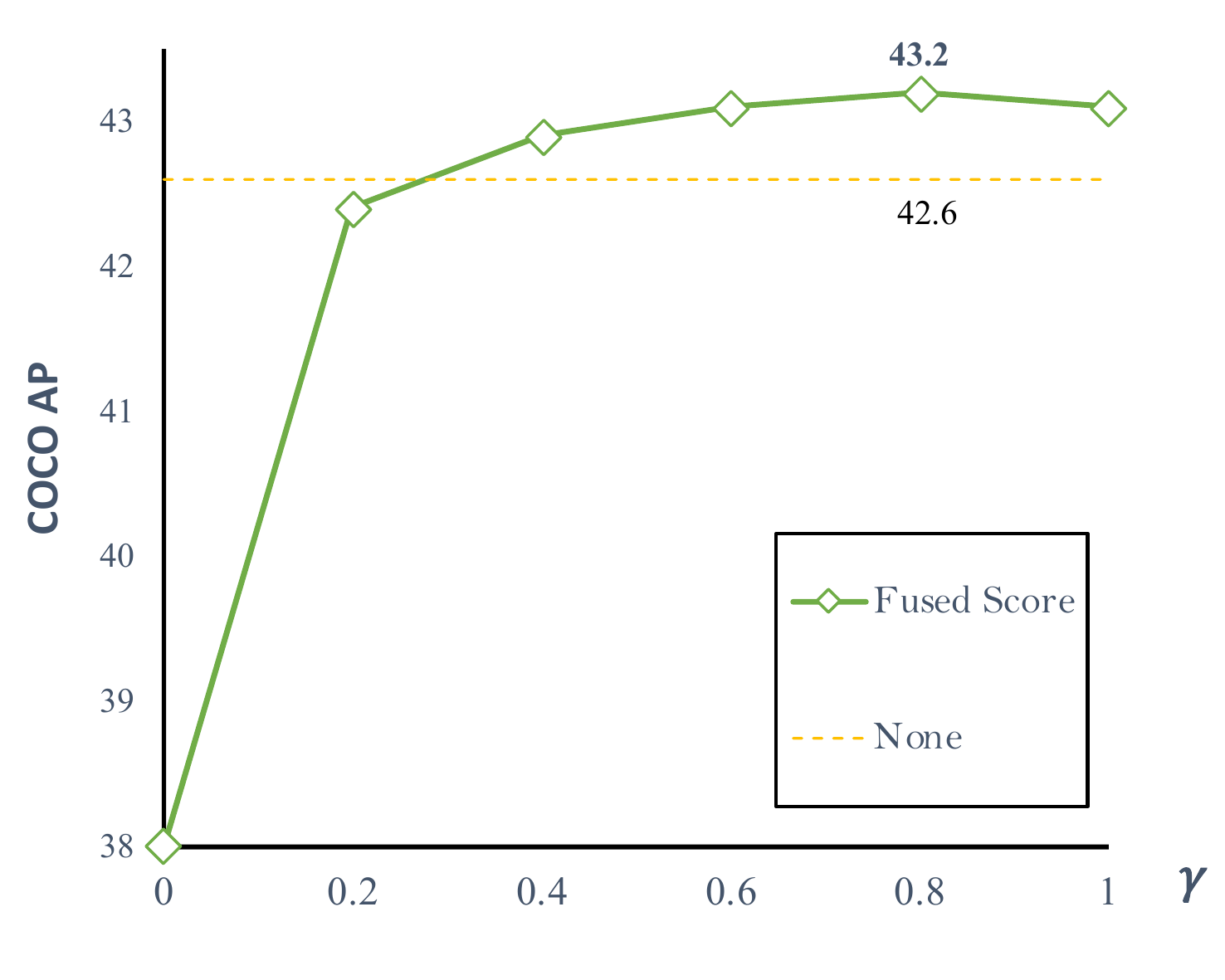}
	\caption{The effectiveness of different balanced factor $\gamma$ on COCO $2017~val$.  ``$None$'' means that the original classification confidence is regarded as score of localization.}
	\label{fig:fused_score}
\end{figure}

For bounding boxes, while our ISM establishes the bond between their spatial quality and final score of localization, we still observe the shortage mentioned in Section~\ref{section:rsm}. As shown in Fig.~\ref{fig:rsm}, although bounding box of incomplete object almost overlaps its ground truth, lower predicted IoU reduces the overall score evidently. Instead, RSM introduces new classification results to lessen disadvantage on incomplete object, and eliminates hysteresis of classification by resampling strategy. In Fig.~\ref{fig:fused_score}, we explore the effectiveness of fused score. The result shows mixing 0.2 of RSM score can further improve the performance. It is worth noting that fused scoring method relatively brings 1.4\% improvement over original scoring way.

\subsection{Comparison with State-of-the-arts}

We compare our CPM R-CNN with other state-of-the-art anchor-based methods on COCO test-dev in Table \ref{table:all}. For multi-scale training, we select a scale between 600 to 800 as the shorter side of images. We set batch size of RoIs to 32 per image when using ResNeXt-64$\times$4d-101 as the backbone. Other experiment settings are consistent with those mentioned in Section~\ref{section:set}.

\setlength{\tabcolsep}{4pt} 
\begin{table*}
\renewcommand\arraystretch{1.2}
\small
\begin{center}
\begin{tabular}{l|c|ccc|ccc}
 \hline
method &  backbone &  AP & AP$_{0.5}$ & AP$_{0.75}$ & AP$_{S}$ & AP$_{M}$ & AP$_{L}$\\
\hline
YOLOv2~\cite{Redmon2017YOLO9000} & DarkNet-19 & 21.6 & 44.0 & 19.2 & 5.0 & 22.4 & 35.5\\
SSD-512~\cite{Liu2016SSD} & ResNet-101 & 31.2 & 50.4 & 33.3 & 10.2 & 34.5 & 49.8\\
DSSD-513~\cite{Fu2017DSSD} & ResNet-101 & 33.2 & 53.3 & 35.2 & 13.0 & 35.4 & 51.1\\
RefineDet512~\cite{Zhang2018Single} & ResNet-101 & 36.4 & 57.5 & 39.5 & 16.6 & 39.9 & 51.4\\
RetinaNet800~\cite{Lin2017Focal} & ResNet-101 & 39.1 & 59.1 & 42.3 & 21.8 & 42.7 & 50.2\\
Cas-RetinaNet800~\cite{zhang2019cascade} & ResNet-101 & 39.3 & 59.0 & 42.8 & 22.4 & 42.6 & 50.0\\
HSD-512~\cite{cao2019hierarchical} & ResNet-101 & 40.2 & 59.4 & 44.0 & 20.0 & 44.4 & 54.9\\
\hline
Faster R-CNN+++~\cite{He2016Deep} & ResNet-101 & 34.9 & 55.7 & 37.4 & 15.6 & 38.7 & 50.9\\
Faster R-CNN w FPN~\cite{Lin2017Feature} & ResNet-101 & 36.2 & 59.1 & 39.0 & 18.2 & 39.0 & 48.2\\
Mask R-CNN w FPN~\cite{He2017Mask} & ResNet-101 & 38.2 & 60.3 & 41.7 & 20.1 & 41.1 & 50.2\\
MS R-CNN w FPN~\cite{Huang2019Mask} & ResNet-101 & 38.3 & 58.8 & 41.5 & 17.8 & 40.4 & 54.4\\
Revisiting R-CNN w FPN~\cite{Cheng2018Revisiting} & ResNet-101\&152 & 40.7 & 64.4 & 44.6 & 24.3 & 43.7 & 51.9\\
RPDet w FPN~\cite{Cai2018Cascade} & ResNet-101 & 41.0 & 62.9 & 44.3 & 23.6 & 44.1 & 51.7\\
Grid R-CNN w FPN~\cite{lu2019grid} & ResNet-101 & 41.5 & 60.9 & 44.5 & 23.3 & 44.9 & 53.1\\ 
Cascade R-CNN w FPN~\cite{yang2019reppoints} & ResNet-101 & 42.8 & 62.1 & 46.3 & 23.7 & 45.5 & 55.2\\
\hline
$ours$ : &&&&&&\\
CPM R-CNN w FPN & ResNet-50 & 41.7 & 59.2 & 44.4 & 23.1 & 44.0 & 54.7\\  
CPM R-CNN w FPN & ResNet-101 & 43.3 & 61.2 & 46.1 & 23.9 & 46.3 & 56.6\\ 
CPM R-CNN w FPN & ResNeXt-64$\times$4d-101 & 43.6 & 62.0 & 46.5 & 24.7 & 46.5 & 57.3\\
CPM R-CNN w FPN & ResNeXt-64$\times$4d-101-DCN & 46.4 & 65.3 & 49.5 & 26.8 & 49.4 & 61.0\\
CPM R-CNN w FPN$^{+}$ & ResNeXt-64$\times$4d-101-DCN & 47.4 & 66.6 & 50.6 & 28.5 & 51.0 & 61.1\\
CPM R-CNN w FPN$^{+*}$ & ResNeXt-64$\times$4d-101-DCN & \bf 49.9 & \bf68.7 & \bf53.4 & \bf31.5 & \bf53.3 & \bf63.8\\
\hline
\end{tabular}
\end{center}
\caption{Comparison with state-of-the-art anchor-based detectors on COCO $test-dev$. Bold fonts indicate the best performance. +: utilizing multi-scale training. *: utilizing other strategy such as soft NMS and multi-scale testing.}
  \label{table:all}
\end{table*}

\begin{figure}
	\centering
	\includegraphics[height=6cm]{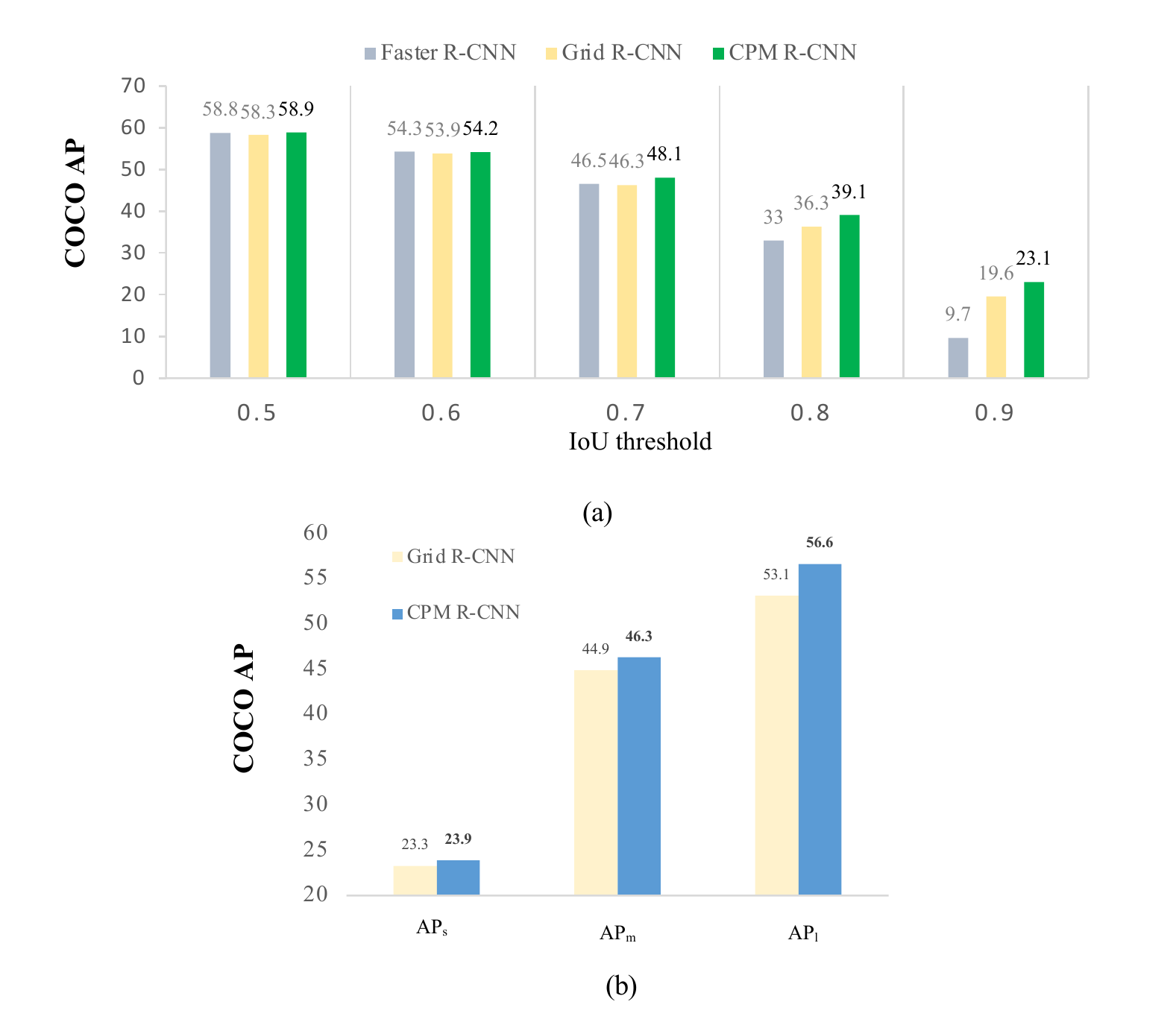}
	\caption{(a): Comparison of AP results across IoU thresholds from 0.5 to 0.9 based on ResNet-50 with FPN. (b): AP results on different scales based on ResNet-101 with FPN.}
	\label{fig:sml}
\end{figure}

\subsubsection{Accuracy  and speed analysis}
Based on ResNet-50, our CPM R-CNN achieves the leading AP performance 41.7 in COCO test-dev without bells-and-whistles, which is even better than Grid R-CNN with ResNet-101.Together with other strategy, we can further achieve 49.9\% AP. In addition, we analyse AP results across different IoU thresholds as shown in Fig.~\ref{fig:sml} (a). Compared with Faster R-CNN and Grid R-CNN, at higher IoU thresholds, our method show a noticeable improvement which are and (6.1\%, 2.8\%) and ($\bf13.4\%, 3.5\%$) in 0.8 and 0.9 respectively. This is attributed to cascade coarse-to-fine optimization from presented CMM. Gradual tightening mapping strategy release the limitation of manual anchors for box regression, especially when box comes from heatmap prediction. In addition, in terms of speed, CPM has great advantage over original Grid R-CNN. Although CPM may be slightly slower compared with plus vision, it has huge contribution in AP.

\subsubsection{Significant performance in large objects}
We also evaluate our method on objects of different scales. In Fig.~\ref{fig:sml} (b), our detector could achieve 46.3\% $AP_{m}$ and 56.6\% $AP_{l}$, which enhance performance in large-scale objects significantly compared to the Grid R-CNN (44.9\%/53.2\%). We guess anchor-based architecture limits distribution of candidate proposals in large-scale objects by the influence of RPN, thus this phenomenon will reduce the integrity of heatmap at second stage. Meanwhile, it uses classification confidence as a criterion of localization quality, but this way likely overestimates their location accuracy for large-scale objects. In CPM R-CNN, three modules can jointly solve the above problems.

\setlength{\tabcolsep}{4pt}
\begin{table}
\renewcommand\arraystretch{1.1}
\small
\begin{center}
\begin{tabular}{c|c|c|ccccccccccccc}
\noalign{\bigskip}
\hline
Method & backbone & AP & speed \\
\hline
Grid R-CNN & R-101-FPN & 41.3 & 0.290 \\
CPM R-CNN (3 stages) & R-101-FPN & 43.2 & 0.140\\
\hline
\end{tabular}
\end{center}
\caption{
The inference speed of each method on TITAN Xp. }
\label{table:resamplescore}
\end{table}
\section{Conclusion}
In this paper, we investigate two major misalignment in anchor-based point-guided method. Our proposed CPM R-CNN is a simple and effective framework, aligning distribution of RoIs and score of localization via cooperation of three unique modules. Specifically, to assist prediction of heatmap to receive required completeness and accuracy in feature extraction, we first optimize matching strategy of original misaligned proposals by applying coarse-to-fine mapping ratio in cascade network structure. Then, in order to realize the accurate estimation of location quality, we supervise the spatial information of bounding boxes by introducing regression of IoU. Besides, hysteresis of scoring in classification is also eliminated by updating candidate samples. In our extensive experiments, we observe this framework shows consistent accuracy gain on MS COCO dataset.

\bibliographystyle{IEEEtran}
\bibliography{egbib}

\end{document}